\documentclass[10pt,twocolumn,letterpaper]{article}

\usepackage{iccv}              
\definecolor{iccvblue}{rgb}{0.21,0.49,0.74}
\usepackage[pagebackref,breaklinks,colorlinks,allcolors=iccvblue]{hyperref}

\usepackage{array} 
\usepackage{booktabs} 
\definecolor{Gray}{gray}{0.9}
\usepackage{lipsum}
\usepackage{amsmath}
\usepackage{tabularx}
\usepackage{multirow}
\usepackage[ruled,linesnumbered]{algorithm2e} 
\usepackage{colortbl}

\def\paperID{11032} 
\def\confName{ICCV}
\def\confYear{2025}

\title{Towards Efficient Vision State Space Models via Token Merging}

\author{Jinyoung Park \quad Minseok Son \quad Changick Kim \quad\\
KAIST\\
{\tt\small \{jinyoungpark,ksos104,changick\}@kaist.ac.kr}
}

\begin{document}
\maketitle
\begin{abstract}
State Space Models (SSMs) have emerged as powerful architectures in computer vision, yet improving their computational efficiency remains crucial for practical and scalable deployment.
While token reduction serves as an effective approach for model efficiency, applying it to SSMs requires careful consideration of their unique sequential modeling capabilities.
In this work, we propose MaMe, a token-merging strategy tailored for SSM-based vision models.
MaMe addresses two key challenges:
quantifying token importance and preserving sequential properties. 
Our approach leverages the state transition parameter $\mathbf{\Delta}$ as an informativeness measure and introduces strategic token arrangements to preserve sequential information flow.
Extensive experiments demonstrate that MaMe achieves superior efficiency-performance trade-offs for both fine-tuned and off-the-shelf models. 
Particularly, our approach maintains robustness even under aggressive token reduction where existing methods undergo significant performance degradation.
Beyond image classification, MaMe shows strong generalization capabilities across video and audio domains, establishing an effective approach for enhancing efficiency in diverse SSM applications.

\end{abstract}
    
\section{Introduction}
\label{sec:intro}

Vision Transformers (ViT) \cite{vit_dosovitskiy} pioneered image processing through patch tokenization, leading other vision models to adopt similar token-based processing \cite{deit_touvron,liu2022videoswin}.
This approach has been successfully extended to State Space Model (SSM)-based architectures, such as Vision Mamba \cite{vim_zhu}.
Vision Mamba achieves notable performance across various vision tasks \cite{videomamba_li, videomamba_park, erol2024aum, hu2024zigma, liang2024pointmamba} by introducing selective scan mechanisms with a hardware-aware design.

As token-based processing continues to advance, improving computational efficiency becomes crucial for practical and scalable deployment.
Among various approaches, reducing the number of tokens emerges as an intuitive solution.
Based on the insight that not all tokens are equally important, early approaches \cite{bat_long,dynamicvit_rao,evit_liang} optimize ViT by simply removing less important tokens, such as those representing background regions or low-level textures.
Recent methods have evolved toward token-merging to minimize information loss.
For instance, some approaches merge similar tokens using attention keys (K) similarity \cite{tome_bolya}, while others leverage attention-based importance scores \cite{mctf_lee,wu2023ppt}.

However, effective token merging in SSM-based vision models remains an open challenge.
Unlike attention-based architectures, SSMs process tokens through state-space equations.
This architectural disparity motivates us to rethink token merging for SSMs, leading us to address two key aspects.
First, SSMs need alternative token importance measure.
Second, as SSMs process information through sequential state updates, it is crucial to carefully handle sequential relationship after merging.
In response, we propose MaMe, a novel token-merging strategy specifically designed for SSM architectures that considers both token importance and sequential order.
Our approach evaluates token importance using $\mathbf{\Delta}$ values naturally derived from state transitions and preserves the sequential relationship by maintaining the original token order.
Extensive experiments demonstrate that MaMe achieves excellent computational efficiency while maintaining competitive performance across various token reduction numbers.
On ImageNet-1K classification, our method even surpasses the base-model performance with reducing computational complexity.
Notably, when applied to ViM models without additional training, MaMe maintains strong performance, whereas the existing approaches suffer from a significant drop.

Furthermore, We extensively analyze MaMe's token-merging behavior with visualization.
Our key contributions are as follows:
\begin{itemize}
    \item  We introduce MaMe, the first token-merging framework specifically designed for SSM-based vision models.
    \item We present a token merge score incorporate a novel informativeness measures from state transition dynamics.
    \item We design a token arrangement that preserves SSM's sequential nature.
    \item  Through extensive experiments, we demonstrate that MaMe achieves superior efficiency-performance trade-offs compared to existing approaches.
\end{itemize}

\section{Related work}
\subsection{State Space Models}
State Space Models (SSMs) built on linear state space equations serve as an alternative to CNNs and Transformers for modeling long-range dependencies.
The S4 model reduces computational bottlenecks through reparameterization, effectively handling long-range dependencies \cite{s4_gu}.
Recently, SSMs have also gained attention in computer vision tasks with achieving performance competitive to previous works \cite{s4d_gu,dss_gupta,liquid-s4_hasani,s5_smith}.
The S4ND model normalizes S4’s parameters into a diagonal structure, becoming the first to achieve linear time complexity, while demonstrating promising results in visual tasks \cite{s4nd_nguyen}.
However, S4-based models inherently focus equally on all elements in the input sequence, regardless of their characteristics.
As a subsequent advancement in SSMs, Mamba \cite{mamba_gu} introduces selective SSM and an efficient hardware-aware design, achieving high performance while maintaining linear computational cost.
This design has inspired multiple studies on its application in visual processing tasks.
ViM \cite{vim_zhu} extends columnar plain ViT's image tokenization process and integrates it with a bidirectional Mamba block, enhancing its ability to handle both position sensitivity and global context.
VMamba \cite{vmamba_liu} introduces four-route scanning on a hierarchical structure to model non-sequential image information.
On the other hand, MambaVision \cite{hatamizadeh2024mambavision} further enhances model performance by introducing a hybrid architecture that strategically combines Mamba blocks with Transformer layers.
Beyond image processing, the reliable performance of Mamba-based models has been validated in various domains, including video \cite{videomamba_park,videomamba_li} and audio \cite{erol2024aum} tasks.

\subsection{Token Reduction}
Based on the premise that not all tokens hold equal importance \cite{evit_liang}, token reduction techniques aim to decrease the number of tokens directly while preserving the performance of foundational models like ViTs, reducing computational costs.
In this context, various token reduction methods have been introduced and evaluated within ViTs, each defining unique criteria for measuring token importance.

Many of works \cite{star_zhang,ats_fayyaz,wu2023ppt} use token-to-token similarity or attention weights to the class token to measure importance and remove tokens accordingly.
ViT-ToGo \cite{vit-togo_lee} estimates token importance for pruning based on attention probabilities from each head.
Approaches like DynamicViT \cite{dynamicvit_rao} and IA-RED2 \cite{iared2_pan}, on the other hand, incorporate additional modules to learn token importance dynamically.
Following the success of token pruning techniques applied to ViT, compatible methods with ViM-based models have also been introduced.
ToP \cite{etp_zhan} 
defines the importance of tokens directly by summing the output tokens of SSM over the channel dimension and pruning the less important ones.
Notably, it maintains state computational flow by updating hidden states even at pruned token positions during the scan process. 
While this technique improves accuracy compared to naive pruning, it has fundamental limitations—the approach still requires SSM operations for the original number of tokens and completely loses information from pruned tokens.

Token pruning, which reduces the computational cost by removing less important tokens, is one effective approach.
However, it inevitably results in some loss of information.
As a result, performance drops significantly with higher pruning ratios, limiting its practical application when substantial computational savings are needed.
Consequently, recent methods have aimed to minimize information loss by employing token merging techniques that enhance model efficiency \cite{evit_liang,dynamicvit_rao,spvit_kong,tokenpolling_marin,tome_bolya,bat_long,mctf_lee, bat_long,tofu_kim,algm_norouzi,vomix_peng,ltm_wang,mctf_lee}.
EViT \cite{evit_liang} uses attentiveness to the [CLS] token as a measure of importance for selecting tokens to merge, while SPViT \cite{spvit_kong} introduces an extra layer to perform this calculation.
A parallel line of study for token selection methods, such as token pooling \cite{tokenpolling_marin} and ToMe \cite{tome_bolya}, leverages the similarity between attention keys (K) to reduce redundancy among tokens.
MCTF \cite{mctf_lee}, one of the recent works, enhances both efficiency and performance by simultaneously considering attention keys (K) similarity, attentiveness to the class token, and size to estimate token importance.

While token merging approaches have shown promise in Transformer-based architectures, applying them to SSM-based vision models presents unique challenges. First, the token importance metrics used in Transformers commonly rely on attention mechanisms that don't exist in SSM models, requiring new approaches to identify which tokens should be merged. Second, unlike attention mechanisms that naturally compute relationships between all tokens, SSMs process tokens sequentially, making the positioning of merged tokens critical for maintaining proper information flow. 
In our work, we address these challenges by proposing a merge score as an importance metric, incorporating novel token informativeness specifically designed for SSMs, and introducing strategic token arrangements 
to minimize information loss.

\begin{figure*}[ht!]
\begin{center}
\includegraphics[width=\linewidth]{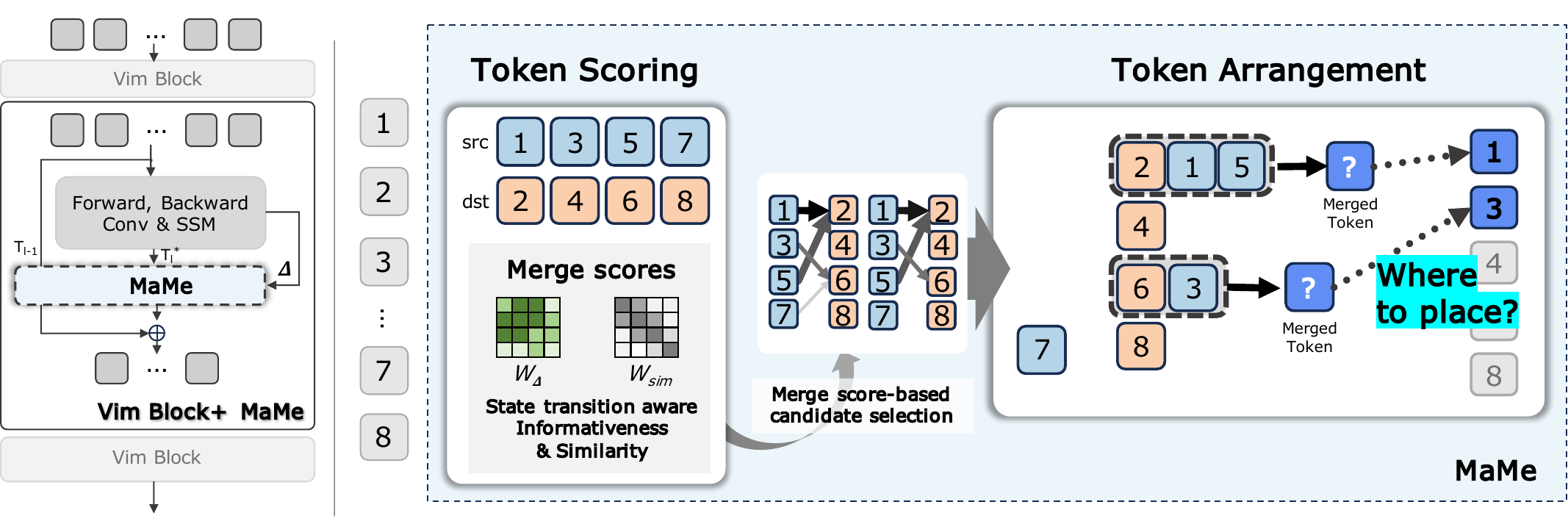}
\end{center}
\caption{Overview of the MaMe framework.
Each MaMe block operates through three sequential phases: scoring, candidate selection, and token arrangement.
}
\label{fig:overall_framework}
\end{figure*}

\section{Methods}
\subsection{Preliminaries}
\subsubsection{State Space Models}
State space models (SSMs) are introduced to model sequential data.
SSM predicts output $y(t) \in \mathbb{R}$ through hidden state $h(t) \in \mathbb{R}^{C}$ computed using input data $x(t) \in \mathbb{R}$ in a current step $t$, which can be mathematically expressed as ordinary differential equations (ODE) as follows:
\begin{gather} \label{eq:ssm_orig}
\begin{aligned}
    h'(t) &= \mathbf{A}h(t) + \mathbf{B}x(t), \\
    y(t) &= \mathbf{C}h(t),
\end{aligned}
\end{gather}
where $\mathbf{A} \in \mathbb{R}^{C \times C}$ represents the evolution matrix for the current hidden state, and $\mathbf{B} \in \mathbb{R}^{1 \times C}$ and $\mathbf{C} \in \mathbb{R}^{1 \times C}$ are the projection matrices for the input and the hidden state, respectively.
\Cref{eq:ssm_orig} is discretized using a zero-order hold (ZOH) technique, which retains the signal for a step size of $\mathbf{\Delta}$, making it suitable for processing discrete signals such as text and images.
The parameter discretization process is formulated as follows:
\begin{gather} \label{eq:zoh}
\begin{aligned}
    \mathbf{\bar{A}} &= \exp(\mathbf{\Delta A}), \\
    \mathbf{\bar{B}} &= (\mathbf{\Delta A})^{-1}(\exp(\mathbf{\Delta A})-\mathbf{I})\cdot\mathbf{\Delta B}.
\end{aligned}
\end{gather}

Finally, the recursive form of the discretized SSM is defined as follows:
\begin{gather} \label{eq:ssm_rec}
\begin{aligned}
    h(t) &= \mathbf{\bar{A}}h(t-1) + \mathbf{\bar{B}}x(t), \\
    y(t) &= \mathbf{C}h(t).
\end{aligned}
\end{gather}
Mamba \cite{mamba_gu} is one of the discrete versions of SSM, implementing a selective scan mechanism (S6).
S6 includes $\mathbf{B}$, $\mathbf{C}$ and $\mathbf{\Delta}$ in \cref{eq:ssm_orig,eq:zoh}, which are determined dynamically based on the input.
This allows the model to better capture the context of the input sequence.

\subsubsection{Vision Mamba} \label{subsubsec:vim}
Our method is designed based on Vision Mamba (ViM) \cite{vim_zhu}.
ViM successfully adapts the Mamba architecture to visual tasks by introducing bidirectional sequence modeling through forward and backward scans.
The bidirectional SSM of ViM can be expressed in a simplified mathematical form as follows:
\begin{gather} \label{eq:vim}
\begin{aligned}
    Y_{d} &\leftarrow SSM(\mathbf{A}_{d},\mathbf{B}_{d},\mathbf{C}_{d})(X_{d}), \quad \text{for } d \in \{f, b\}, 
    \\
    T_{l} &\leftarrow T^{*}_{l}+T_{l-1},
    \\
    T^{*}_{l} &= Linear(Y_{f}+Y_{b}),
\end{aligned}
\end{gather}
where $X \in \mathbb{R}^{B \times N \times C}$ represents the projected input token sequence, $Y \in \mathbb{R}^{B \times N \times C}$ is the output from the SSM corresponding to $X$, and $d$ denotes the modeling direction, which can be either forward $f$ or backward $b$.
$B$ and $N$ denotes a batch size and number of tokens, respectively.
A token $T_{l} \in \mathbb{R}^{B \times N \times D}$ in a current layer $l$ is updated as the projected sum of the previous $T_{l-1}$ and $Y$.

\subsection{MaMe}
The main goal of MaMe is to reduce the number of tokens in the input sequence $T \in \mathbb{R}^{N \times D}$ to obtain an output sequence $T' \in \mathbb{R}^{(N-r) \times D}$, where $r$ is the number of reduced tokens, while minimally interfering with the information in $T$ and the relationships between tokens within the sequence.
To achieve this, we utilize the intermediate output token $T^{*}_l$ of SSM block in a $l$-th layer.
To compute the final reduced output token using the decreased token set $T^{*'}_{l}$, tokens with the same indices in $T^{*'}_{l}$ are merged in $T_{l-1}$, and part of equation (4) is modified as follows:
\begin{equation} \label{eq:vim_merge}
    T'_{l} \leftarrow T^{*'}_{l}+T'_{l-1}.
\end{equation}

In MaMe, two primary components are proposed: (1) a token merge score and (2) a token arrangement.
The overall framework of our proposed methods is depicted in \cref{fig:overall_framework}.
The MaMe block is placed right after the SSM block at token merging layers, reducing the number of tokens from the current step.
We set three of layers as the token merging layers, and the indices of the layers are defined in \cref{subsec:exp_setting}.

\noindent\textbf{Token Merge Score.}
We first aim to reduce redundant contextual information.
To achieve this, we utilize $T^{*}_l$, which represents an result of sequential modeling at $l$-th layer of SSM.
Each output token contains both unique information of itself and its sequential relationship with other tokens.
To minimize unnecessary redundancy, we calculate the similarity between the $i$-th and $j$-th output tokens using the following equation:

\begin{equation}
     W_{\text{sim}}(T^{*i}_{l}, T^{*j}_{l}) = \frac{1}{2} \cdot \left(\frac{T^{*i}_{l} \cdot T^{*j}_{l}}{\lVert T^{*i}_{l}\rVert \lVert T^{*j}_{l}\rVert} + 1\right).
\end{equation}

While merging tokens based on similarity is intuitive, it often over-merges informative tokens \cite{mctf_lee}, leading to information loss. 
Unlike Transformers, where attention scores on a class token are well-established indicators of token informativeness, SSMs lack such principled informativeness measures. 
To bridge this gap, we propose utilizing $\mathbf{\Delta}$ from SSM as a measure of token informativeness.
To show how $\mathbf{\Delta}$ naturally serves as an informativeness measure, we first examine its role in state transitions. 
By substituting the discretization parameters from \cref{eq:zoh} into \cref{eq:ssm_rec}, we can express the discretized state transition as:
\begin{equation} \label{eq:w_delta}
    h_t = e^{\mathbf{\Delta} \mathbf{A}}h_{t-1} + (\mathbf{\Delta} \mathbf{A})^{-1}(e^{\mathbf{\Delta} \mathbf{A}} - I)\mathbf{\Delta} \mathbf{B} x_t
\end{equation}

This formulation reveals how $\mathbf{\Delta}$ controls information flow in the model. 

Following the standard SSM analysis \cite{mamba_gu, hippo_gu}, $\mathbf{A}$ is defined as a negative integer, such that larger $\mathbf{\Delta}$ emphasizes the current input ($e^{\mathbf{\Delta} \mathbf{A}} \to 0$), while smaller $\mathbf{\Delta}$ preserves the previous state ($e^{\mathbf{\Delta} \mathbf{A}} \to 1$).
This suggests that $\mathbf{\Delta}$ can serve as an indicator of how much new information each token contributes to the state updates.

Indeed, our empirical observations validate this property: regions with high $\mathbf{\Delta}$ values correspond to contextually rich image regions (\cref{subsec:analysis}).

To leverage $\mathbf{\Delta}$ as a token importance measure, we define $\mathbf{\hat{\Delta}} = f(\mathbf{\Delta}_{f},\mathbf{\Delta}_{b})$ in the bidirectional scan of ViM, where $\mathbf{\Delta}_f$ and $\mathbf{\Delta}_b$ represent $\mathbf{\Delta}$ from forward and backward passes, respectively.
$f(\cdot,\cdot)$ indicates a integration function, e.g., max and average.
We experimented with diverse integration functions and selected carefully based on the results in \cref{subsec:ablation}.
To identify token pairs for merging, we calculate the average $\mathbf{\hat{\Delta}}$ value for each token pair as follows:
\begin{equation} \label{eq:w_delta}
    \mathbf{\hat{\Delta}}^{(i,j)}_{l} = \frac{\mathbf{\hat{\Delta}}^{i}_{l}+\mathbf{\hat{\Delta}}^{j}_{l}}{2},
\end{equation}
where $\mathbf{\hat{\Delta}}^{i}_{l}$ represents $\mathbf{\hat{\Delta}}$ of the $i$-th token in $l$-th layer.
The equation for token merge weight based on token informativeness is as follows:
\begin{equation} \label{eq:w_delta}
    W_\mathbf{\Delta} = \exp(-\mathbf{\hat{\Delta}}^{(i,j)}_{l}/\tau),
\end{equation}

where $\tau$ represents a hyperparameter adjusting the impact of $\mathbf{\Delta}$ on the token merge score.

The final token merge score is defined as:
\begin{equation}    \label{eq:tm_score}
    \text{Score} = W_{\text{sim}} \cdot W_\mathbf{\Delta} ,
\end{equation}
with token pairs having higher scores being merged.
Thus, \Cref{eq:w_delta,eq:tm_score} ensure that more informative token pairs are not merged even if their similarity is high.

\noindent\textbf{\\Token Arrangement.}
Given the token merge score defined above, our MaMe adopts a bipartite soft matching \cite{tome_bolya,mctf_lee} to identify candidate token pairs for merging.

This matching algorithm first splits tokens as two sets, i.e., source (src) and destination (dst).
Each token in the source set is matched with the token in the destination set that has the highest token merging score.
After all, the $r$ token pairs are merged based on the destination token.

Following token merging via bipartite matching, the optimal placement of merged tokens emerges as a key consideration. 
State Space Models (SSMs) fundamentally operate through sequential processing, where each token contributes to an evolving hidden state representation that encodes contextual information.
Token merging inherently introduces structural discontinuities in the sequence, as original tokens are removed and their information compressed into merged representations. 
These merged tokens contain information from multiple source tokens—including non-adjacent tokens or previously merged tokens.
Therefore, the positional arrangement of them within the sequence potentially influences state transition dynamics and, consequently, the model's representational capacity.
To systematically address this placement problem, we explore several strategies for token arrangement:
\textbf{(1) Isolated Placement}:
Merged tokens are placed independently at the beginning or end (boundaries) of the sequence, thereby preserving the structure of the unmerged tokens.
\textbf{(2) Internal Insertion}:
Internal insertion merging includes strategies for placing merged tokens at various positions within the sequence:
\textbf{(2.1) Destination Position}:
Merged token candidates are selected by using the destination token as an anchor and then inserted at the destination token’s position.
\textbf{(2.2) Informativeness-based Insertion}:
This method prioritizes the insertion of tokens based on their informativeness. It evaluates the informativeness score of the tokens being merged and selects the optimal position accordingly.
\textbf{(2.3) Order-based Insertion}:
This approach uses either the position of the frontmost or middle position among those being merged as the insertion location.
This method aims to maintain sequential continuity as much as possible.
The detailed explanations for each component are provided in the supplemental material.

\begin{figure*}[ht]
    \centering
    \includegraphics[width=\linewidth]{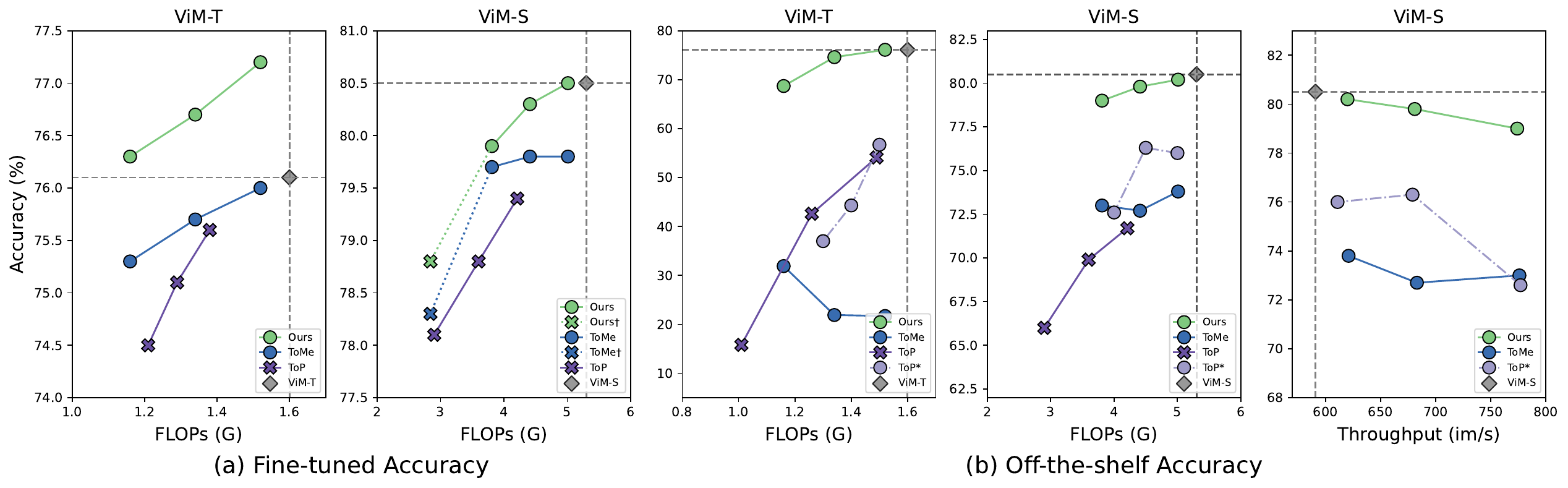} 
    \caption{
    Image classification results under varying FLOPs and throughputs. 
    (a) fine-tuned accuracy and (b) off-the-shelf accuracy on ViM-T and ViM-S models.
    Markers indicate: O - our token reduction configurations with reduction number $r \in{\{10,30,50}\}$; 
    X - ToP \cite{etp_zhan} configurations; 
    ToP* - ToP adapted with our configuration;
    Diamond - base-model without token reduction;
    Our approach achieves superior efficiency-performance trade-offs for both fine-tuned and off-the-shelf models, maintaining robustness even under aggressive token reduction where existing methods undergo significant performance degradation.
    }
    \label{fig:zs}
\end{figure*}

\section{Experiments}

\subsection{Experimental Settings}  \label{subsec:exp_setting}
We evaluated MaMe on ImageNet-1K using ViM and MambaVision, fine-tuning with ImageNet-1K pre-trained models for 30 epochs. 
 For ViM, we applied token merging at layers 8, 14, and 20 with reduction ratio $r=50$. 
MambaVision implementations merged tokens at the initial layer of stage 3 with $r=40$, following swin \cite{dynamicvit_rao}.
ToMe, a similarity-only variant serves as our baseline.
Ablations use ViM-T with $r=50$ and $\tau=10$, and we report top-1 accuracy (\%) and FLOPs (G).
More details are provided in supplementary materials.

\subsection{Experimental Results}
\begin{table}[t]
\centering
\begin{tabular}{p{0.32\columnwidth} >{\centering\arraybackslash}p{0.13\columnwidth} >
{\centering\arraybackslash}p{0.2\columnwidth} >{\centering\arraybackslash}p{0.15\columnwidth}}

\toprule
Method               & Params.
(M)
 & Top-1 Acc. 
 (\%)  & FLOPs
 (G)  \\ 

\midrule
\rowcolor{Gray}
ViM-T \cite{vim_zhu}                   & 7             & 76.1 (-)  & 1.6    \\

 \dag ViM-T \cite{vim_zhu}          & 7            & 76.1 (-)  & 1.5     \\
 
+ \dag EViT~\cite{evit_liang}     & 7            & 71.3 (-4.8)   & 1.3        \\
+ \dag ToP~\cite{etp_zhan}           & 7              & 75.1 (-1.0) & 1.3        \\
+ ToMe~\cite{tome_bolya}        & 7             &75.3 (-0.8)     & 1.2 \\
+ MaMe& 7             & 76.3 (+0.2)     & 1.2       \\
\midrule
\rowcolor{Gray}
ViM-S \cite{vim_zhu}                   & 26        & 80.5  (-)   & 5.3          \\

 \dag ViM-S \cite{vim_zhu}                   & 26               & 80.5  (-)   & 5.1   \\
+ \dag EViT~\cite{evit_liang}     & 26             & 74.8  (-5.7)     & 3.6   \\

+ \dag ToP~\cite{etp_zhan}           & 26             & 78.8  (-1.7)   & 3.6    \\

+ ToMe~\cite{tome_bolya}        & 26             &79.7 (-0.8)     &3.8 \\
+ MaMe  & 26         & 79.9 (-0.6)    & 3.8      \\
\bottomrule
\rowcolor{Gray}
MambaVis.-T \cite{hatamizadeh2024mambavision}              &32     & 82.3 (-)     &4.4      \\
+ ToMe~\cite{tome_bolya}    &32        & 81.4 (-0.9)       & 4.1    \\
+ MaMe                    &  32        & 81.8 (-0.5)    & 4.1\\
\rowcolor{Gray}
MambaVis.-S \cite{hatamizadeh2024mambavision}              & 50	         & 83.3 (-)        & 7.5   \\
+ ToMe~\cite{tome_bolya}    & 50              & 82.2 (-1.1)      & 7.1   \\
+ MaMe                   & 50               &   82.6 (-0.7)     & 7.1  \\

\bottomrule
\end{tabular}
\caption{
Performance comparison on ImageNet-1K. '\dag' indicates results reported by Zhan~\etal\cite{etp_zhan}.
}
\label{tab:vim}
\end{table}

\begin{figure*}[t!]
\begin{center}
\includegraphics[width=2\columnwidth,trim=0 20 0 0]{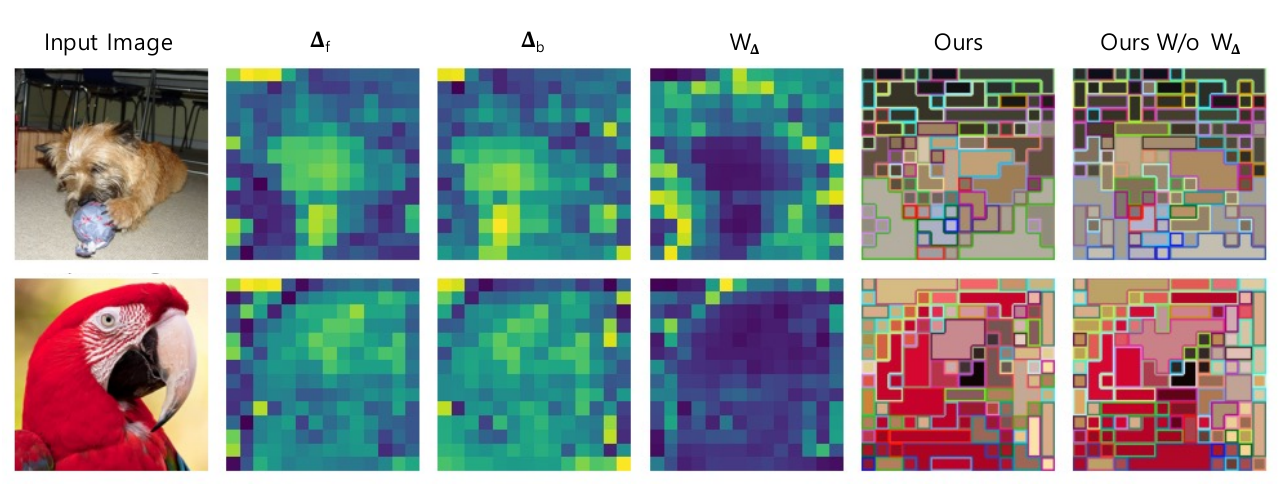} 
\end{center}
\caption{
Token infromative metrics and merging results. 
 From left to right: input images, forward delta ($\mathbf{\Delta}_f$), backward delta ($\mathbf{\Delta}_b$), combined importance weights ($W_\mathbf{\Delta}$), 
 our method (with weighting), and ablation (without weighting). 
  Brighter colors indicate higher values in heatmaps (columns 2-4). 
  Results highlight how informativeness-aware merging selectively preserves semantically relevant regions.
}

\label{fig:vis}
\end{figure*}

 \begin{figure}[t]
 \begin{center}
 \includegraphics[width=\columnwidth,trim=0 20 0 20]{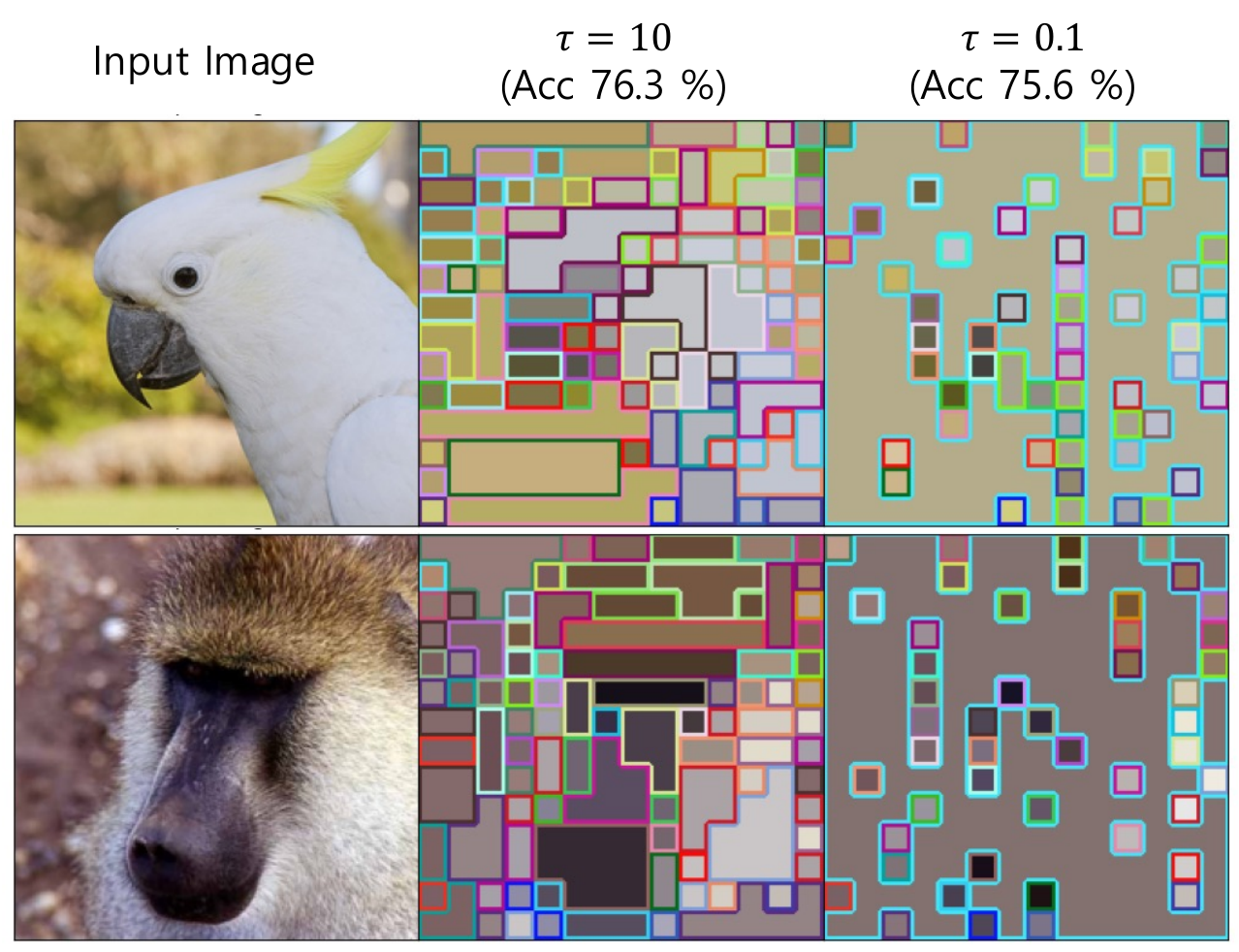}
 \end{center}
 \caption{
Token merging visualizations across different scaling factors $\tau$, with each row representing results from the same input image.
}
 \label{fig:tau_img}
 \end{figure}

We evaluate MaMe against previous token reduction methods across various Mamba-based vision architectures (\cref{tab:vim}). 
Note that our FLOPs calculations use fvcore (details in supplementary material), revealing a higher baseline computational cost than previously reported \cite{etp_zhan}.
MaMe demonstrates consistent performance advantages while maintaining equivalent computational efficiency when applied to both pure SSM architectures (ViM \cite{vim_zhu}) and hybrid models (MambaVision\cite{hatamizadeh2024mambavision}). 
Significantly, while existing token reduction methods typically result in accuracy degradation for ViM-T, MaMe achieves a +0.2\% accuracy improvement despite reducing computational cost by 25\%.

\noindent\textbf{Number of Merged Tokens $r$.}

All token reduction approaches, including token merging, must balance performance preservation against computational savings.
This balance becomes increasingly difficult to maintain with higher token reduction numbers, leading to accelerated accuracy decline.
To examine this trade-off, we designed an experiment comparing existing approaches and our proposed method.
\Cref{fig:zs} (a) shows the accuracy-FLOPs trade-off based across different token reduction techniques.
Each node in the graph represents results using different $r$ values, with $r \in \{10, 30, 50\}$.
The results demonstrate that MaMe maintains higher accuracy as $r$ increases, showing significantly less performance degradation compared to previous methods. 
Particularly at $r=50$ (which removes 75\% of tokens in 224 resolution images), our approach maintains robustness even under this aggressive token reduction, where existing methods undergo significant performance degradation.
Remarkably, for the ViM-T model, our approach actually improves upon the base-model performance.
In our ViM-S experiments, we implemented token reduction at the same layers as ToP to ensure fair comparison (Ours$\dagger$).
When compared under equivalent constraints, MaMe consistently delivers the highest accuracy.

\noindent\textbf{Token Merge without Training.}

MaMe, our proposed method, involves only training-free token-merging operations based on similarity and arrangement criteria.
This approach can be easily integrated into an off-the-shelf model without any additional training.
\Cref{fig:zs} (b) demonstrates the effectiveness of our proposed method when applied to an off-the-shelf model without fine-tuning.
Unlike existing techniques that significantly degrade performance in the same configuration as MaMe, our method preserves high performance.
The graph in the third column of \cref{fig:zs} (b) compares performance across configurations that yield similar throughput for each experimental result.
These results confirm that MaMe delivers superior efficiency-performance balance.

\subsection{Analysis}   \label{subsec:analysis}

\noindent\textbf{Delta as Informativeness Measures.}

In this section, we conduct an in-depth analysis through various visualizations to understand the role and contribution of the token merge score with $\mathbf{\Delta}$ in our framework.

\Cref{fig:vis} illustrates the $\mathbf{\Delta}$ and weight matrix $W_\mathbf{\Delta}$ for each token, resulting in merged tokens at the final merging layer.

Both forward and backward informativeness measures ($\mathbf{\Delta}_f$ and $\mathbf{\Delta}_b$) consistently assign higher values to image foreground regions, with minor variations between them. 
The derived weight matrix $W_\mathbf{\Delta}$ effectively quantifies token informativeness, serving as a key component in our scoring method.
Columns 5-6 demonstrate the practical impact of our approach compared to similarity-only merging. 
Our method strategically preserves foreground tokens while merging background regions. 
In contrast, similarity-only merging often consolidates informative regions (e.g., the dog's face in the first example).
By incorporating both similarity and informativeness, our approach preferentially preserves semantically significant tokens, maintaining representation quality.

\noindent\textbf{\\The Impact of Delta Scaling.}
We investigate how temperature ($\tau$) affects token merging across our MaMe framework.
Our formulation (\cref{eq:w_delta}) uses temperature $\tau$ to control the balance between semantic similarity and informativeness when determining which tokens to merge.
High $\tau$ values prioritize token similarity, potentially merging tokens that play crucial roles for state updates in SSMs.
At low $\tau$ values ($\tau=0.1$), the model aggressively merges tokens based primarily on informativeness, disregarding semantic boundaries and leading to degraded visual coherence. (\cref{fig:tau_img}, 3rd column). 
\Cref{fig:tau_img} shows that setting $\tau$ to 10 creates a balanced consideration of both token similarity and informativeness.
This balance enables merging similar tokens while effectively preserving the original information.
Quantitatively, model accuracy peaks at $\tau=10$ (76.3\%) and drops at higher temperature values (\cref{tab:tau}).
Based on these findings, we set $\tau=10$ for all subsequent experiments.


\begin{table}[t]
\centering
\begin{tabular}{p{0.18\columnwidth} 
>{\centering\arraybackslash}p{0.14\columnwidth} 
>{\centering\arraybackslash}p{0.14\columnwidth} 
>{\centering\arraybackslash}p{0.14\columnwidth} 
>{\centering\arraybackslash}p{0.1\columnwidth}
} 
\toprule

 Method& Acc. (\%) & w/o ft. (\%) & FLOPs (G) & im/s  \\ 
\midrule

ViM-T  &\multicolumn{2}{c}{76.1}  & 1.60  & 1378 \\  
\midrule

Random  &   73.8&10.1 &1.16&1699\\
\midrule
Iso. last   &   75.6 &36.1&1.16&1694\\
Iso. front   &   75.6 & 13.6&1.16&1695\\
\midrule

dst.    & 76.2& 66.8 & 1.16 &  1697  \\
Inform.& 76.2& 65.5 & 1.16 &   1694 \\
Ord. mid    & 76.0 &64.6 & 1.16 & 1695   \\
\rowcolor{Gray}
Ord. front  & 76.3 &68.5 & 1.16 & 1696  \\
\bottomrule
\end{tabular}
\caption{Token Arrangement performance.
Comparison of different token arrangement strategies for merged tokens on ViM-T with fine-tuning accuracy (Acc, \%) and off-the-shelf model accuracy (w/o ft.\%).
}
\label{tab:token_arrangement}
\end{table}

\noindent\textbf{Token Arrangement.}
\Cref{tab:token_arrangement} presents a comparison of token arrangement strategies following token merging. 
All methods maintain consistent computational efficiency (1.16 FLOPs) and inference speed ($\sim$1695 img/s). 
The frontmost order position achieves the highest accuracy (76.3\%), outperforming both the baseline ViM-T model (76.1\%) and random placement (73.8\%). 
Performance without fine-tuning reveals significant differences between strategies, with the frontmost order position maintaining 68.5\% accuracy. 
These results suggest that preserving sequential structure through strategic token positioning is important for SSM-based models, with frontmost order positioning of merged tokens being particularly effective.

\subsection{Ablation Study} \label{subsec:ablation}


\noindent\textbf{Contributions of MaMe Components.}
 Our method achieves 76.3\% (+0.2\%$\uparrow$) accuracy while maintaining efficient computation (1.16 FLOPs) and fast inference speed (1695 img/s). 
Both $\mathbf{\Delta}$-based weighting and token arrangement introduce negligible computation cost while providing significant accuracy gains. 
These components provide a 1\% performance gain over the similarity-only method, effectively optimizing the accuracy-efficiency trade-off.

\begin{table}[t]
\centering
\begin{tabular}{
>{\centering\arraybackslash}p{0.05\columnwidth} 
>{\centering\arraybackslash}p{0.05\columnwidth} 
>{\centering\arraybackslash}p{0.05\columnwidth} 
>{\centering\arraybackslash}p{0.12\columnwidth} 
>{\centering\arraybackslash}p{0.12\columnwidth}
>{\centering\arraybackslash}p{0.12\columnwidth}
>{\centering\arraybackslash}p{0.12\columnwidth}
}
\toprule
sim. & $\mathbf{\Delta}$ & order & Acc. (\%) & FLOPs (G) &im/s& speedup\\
\midrule
& & & \textcolor{gray}{76.1} & \textcolor{gray}{1.60}&\textcolor{gray}{1378}&\textcolor{gray}{$\times$1}\\
 & & & 74.4 & 1.16&1708 &$\times$1.24\\
\checkmark & & & 75.3 & 1.16& 1699&$\times$1.23\\
\checkmark & \checkmark & & 75.8 & 1.16 &1698&$\times$1.23\\
\checkmark & & \checkmark & 76.1 & 1.16& 1697&$\times$1.23\\
\rowcolor{Gray}
\checkmark & \checkmark & \checkmark & 76.3 & 1.16 &1695&$\times$1.23\\
\bottomrule
\end{tabular}
\caption{
Component analysis of MaMe. sim. and $\mathbf{\Delta}$ represents $W_{sim}$ and $W_\mathbf{\Delta}$ in token merge score, respectively, and order denotes token arrangement.
Random score assignment with bipartite matching is reported for all components disabled.
The sim-only case is equivalent to ToMe \cite{tome_bolya}.
}
\label{tab:abl_components}
\end{table}

\begin{table}[t]
  \begin{minipage}[t]{0.43\columnwidth}
    \centering
    \begin{tabular}{
      p{0.22\linewidth}
      >{\centering\arraybackslash}p{0.42\linewidth}}
      \toprule
      $\tau$ & Acc. (\%) \\ 
      \midrule
      0.1 & 75.6 \\
      1   & 76.0 \\
      \rowcolor{Gray}
      10  & 76.3 \\
      20  & 76.2 \\
      50  & 76.1 \\
      \bottomrule
    \end{tabular}
    \captionof{table}{Scaling factor $\tau$.}
    \label{tab:tau}
  \end{minipage}
  \hfill
  \begin{minipage}[t]{0.53\columnwidth}
    \centering
    \begin{tabular}{
      p{0.1\linewidth}
      >{\centering\arraybackslash}p{0.3\linewidth}
      >{\centering\arraybackslash}p{0.36\linewidth}}
      \toprule
      $f$ & Acc. (\%) & FLOPs (G) \\ 
      \midrule
      Max & 76.2 & 1.16 \\
      Min & 76.1 & 1.16 \\
      \rowcolor{Gray}
      Avg & 76.2 & 1.16 \\
      Sum & 76.1 & 1.16 \\
      \bottomrule
      &&\\
      
    \end{tabular}
    \captionof{table}{Integrate function $f$.}
    \label{tab:delta_int}
  \end{minipage}
\end{table}


\noindent\textbf{Delta Integration Function $f$.}
ViM’s bidirectional scanning mechanism calculates both forward and backward $\mathbf{\Delta}$ for each token, capturing distinct information based on the direction of sequential modeling.
To leverage these differing $\mathbf{\Delta}$, we experimented with various integration functions.
In \cref{tab:delta_int}, ``Max" and ``Min" denote criteria for selecting one $\mathbf{\Delta}$ from the two directional $\mathbf{\Delta}$ for each token.
Results show marginal performance differences across integration functions, leading us to adopt the ``average" function to retain information from both directional $\mathbf{\Delta}$.

\newcolumntype{Y}{>{\centering\arraybackslash}X}
\begin{table}[t]
\centering
\begin{tabularx}{0.48\textwidth}{c|c|YY}
\toprule
 & Indices & Acc. (\%) & FLOPs (G) \\
\midrule
ViM-T    &-& 76.1 & 1.60 \\
\midrule
\rowcolor{Gray}
Ours    & [8, 14, 20] & 76.3 & 1.16 \\
\midrule
Shallow                    & [4, 12, 18]      & 75.6 & 1.03 \\
Deep                       & [12, 16, 22]     & 76.7 & 1.29 \\
Even                         & [6, 12, 18]      & 75.9 & 1.06 \\
\midrule
\multirow{2}{*}{4 layers}    & [8, 12, 16, 20]  & 76.1 & 1.15 \\
                             & [6, 12, 18, 22]  & 76.4 & 1.18 \\
\midrule
7 layers &          [8, 10, 12, 14, 16, 18, 20] & 76.1 & 1.15  \\
\midrule
Every layer                &  [0, 1, 2, $\cdots$ , 21, 22, 23]     & 75.3 & 1.12 \\
\midrule
ToP layer                &  [10, 20]     & 76.8 & 1.34 \\                           

\bottomrule
\end{tabularx}
\caption{
Token merging layer selection. 
All configurations reduce the total token count by approximately 150 tokens.
Indices indicate merging layer positions.
}
\label{tab:merge_idx}
\end{table}

\noindent\textbf{\\Token Merging Layer Placement.}
\label{exp: merge locations}
\Cref{tab:merge_idx} examines token merging layer placement in the 24-layer ViM-T model. 
Our standard configuration places merging at the 8th, 14th, and 20th layers (approximately 1/3, 3/5, and 5/6 network depth), following proportions established in prior token-reduction works \cite{evit_liang,dynamicvit_rao}.
We use this same configuration for both ViM-T and ViM-S models (24 layers).
We observe clear trade-offs across different placements. Deeper positioning (12th, 16th, 22nd layers) improves accuracy (76.7\%) but with higher FLOPs (1.29G). 
Earlier positioning (4th, 12th, 18th layers) further reduces FLOPs (1.03G) but sacrifices accuracy (75.6\%).
We tested additional configurations, including four-layer and seven-layer merging, which maintained base-model accuracy but with diminished efficiency gains. Merging at every layer degraded performance (75.3\%), while the ToP layer approach (10th, 20th layers) achieves the highest accuracy (76.8\%) with moderate computational savings.
Based on these findings, we selected the 8th, 14th, and 20th layers as our standard configuration, balancing accuracy and computational efficiency.

\section{Results on Extended Applications}
\subsection{Video Recognition} 

Video processing presents substantial computational challenges due to the additional temporal dimension, making efficient token handling particularly valuable.
Table 7 shows results on the Something-Something V2 dataset using VideoMamba \citep{videomamba_park}. 
Following established fine-tuning protocols \cite{videomamba_park}, we initialized with ImageNet-1K pre-trained weights and train for 35 epochs.
\textbf{Results.}
Our MaMe approach with $r=300$ improves accuracy by 0.1\% (63.8\% vs 63.7\%) while reducing FLOPs from 43 to 34G and increasing throughput by 26\% (97.2 vs 77.2 clips/s). 
Increasing to $r=350$ maintains baseline accuracy while further reducing FLOPs to 32G and achieving 100.1 clips/s throughput.
Notably, MaMe outperforms ToMe when applied to the same model, with ToMe showing a 0.5\% accuracy drop (63.2\%) at the same computational cost. 
These results demonstrate that our approach effectively reduces redundancy in spatiotemporal token representations, making it particularly well-suited for computationally intensive video understanding tasks.

\begin{table}[t]
\centering
\begin{tabular}{
p{0.31\columnwidth}>
{\centering\arraybackslash}p{0.15\columnwidth}>
{\centering\arraybackslash}p{0.08\columnwidth}>
{\centering\arraybackslash}p{0.1\columnwidth}>
{\centering\arraybackslash}p{0.08\columnwidth}}

\toprule
Method               & input    & Acc. (\%)  & FLOPs (G) & clip/s \\ 
\midrule
VideoSwin-T \cite{liu2022videoswin}         & 32×224$^2$ &  52.3      &  75 & 54.2  \\

\midrule
VideoMamba \cite{videomamba_park}   &  16×224$^2$  & 63.7      &  43  & 77.2 \\
+ ToMe$_{r=300}$     &  16×224$^2$ &   63.2     & 34  & 98.2    \\
+ MaMe$_{r=300}$     &  16×224$^2$ &  63.8     & 34   & 97.2     \\
+ MaMe$_{r=350}$     &  16×224$^2$ &  63.7     & 32   & 100.1   \\

\bottomrule
\end{tabular}
\caption{Video classification results on Something-Something V2.}
\label{tab:video}
\end{table}

\begin{table}[t]
\centering
\begin{tabular}
{
p{0.3\columnwidth}>
{\centering\arraybackslash}p{0.2\columnwidth}>
{\centering\arraybackslash}p{0.25\columnwidth}
}
\toprule
Method         &Acc.  (\%) & FLOPs (G)   \\
\midrule
AST-S \cite{gong21b_ast}         & 97.38 & 1.43\\
\midrule

AuM-S \cite{erol2024aum} &  97.51 &  1.73 \\
+ MaMe$_{r=12}$          &  97.59 &  1.37 \\
+ MaMe$_{r=15}$          &  97.42 &  1.28 \\

\bottomrule
\end{tabular}
\caption{Audio classification results on Speech Commands V2.}
\label{tab:audio}
\end{table}

\subsection{Audio Classification } 
Audio can also benefit from efficient token handling, as demonstrated by our evaluation using the Speech Commands V2 dataset \cite{warden2018speech}, which contains approximately 105,000 one-second audio recordings covering 35 common speech commands.
Since AuM \cite{erol2024aum} processes audio by converting signals into spectrogram images, our token merging approach integrates easily. 
Following standard practices \cite{gong21b_ast,erol2024aum}, we initialized models with ImageNet-1K pre-trained weights and report top-1 accuracy.
\textbf{Results.}
Our method demonstrates clear efficiency-performance trade-offs in audio processing. 
MaMe$_{r=6}$ outperforms AuM-S by achieving higher accuracy with reduced computational load. 
Even with more aggressive merging, MaMe$_{r=12}$ still surpasses AST-S in both performance and efficiency.
These results further validate the versatility of our approach across modalities, showing effectiveness for both video and audio data when processed through state space models.

\section{Conclusion}
In this work, we presented MaMe, a token merging strategy for efficient SSM-based vision models. 
We exploited $\mathbf{\Delta}$ from SSM as a token informativeness measure and evaluated its effectiveness, while also explored token arrangement suitable for the model's sequential nature.
Through these efforts, MaMe achieves significant computational benefits while preserving model capabilities.
Comprehensive and analytical experiments and extension to video and audio domains demonstrate the broad applicability of our method, indicating that careful token selection and merging can serve as a practical pathway toward more efficient computation of SSM architectures.

{
    \small
    \bibliographystyle{ieeenat_fullname}
    \bibliography{main}
}

\clearpage
\setcounter{page}{1}
\maketitlesupplementary

\renewcommand{\thesection}{\Alph{section}}
\setcounter{section}{0}
\setcounter{page}{0}

In this supplementary material, we first present additional details in Sec.\ref{sec:imples}.
Then, we provide extended experimental results in Sec.\ref{Extended}, including 1) \textbf{\textit{without training}} \ref{exp: without training}, and 2) \textbf{\textit{finetuning on high-resolution images}} \ref{exp: higher-resolution}.
Finally, we present additional visualizations in Sec.\ref{exp: vis}
\section{Additional Details}
\label{sec:imples}
\subsection{Token Arrangement}
\begin{figure}[t!]
\includegraphics[width=\linewidth]{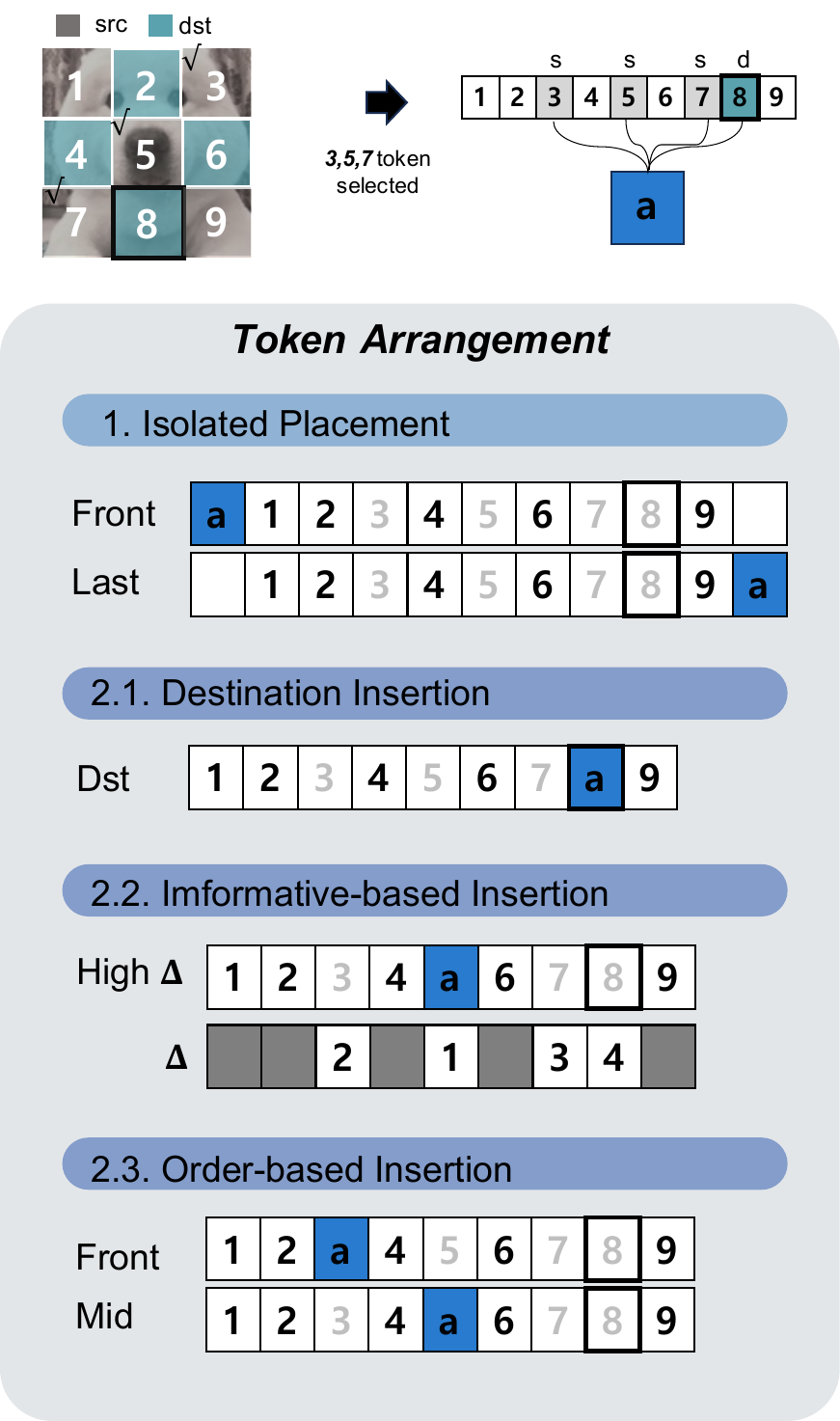}
\caption{Token arrangement strategies following merging operations. 
We illustrate various placement methods for the merged token 'a' (combining tokens 3, 5, 7, and the destination token 8): isolated placement at sequence boundaries (front/last), destination-based positioning, importance-weighted ($\mathbf{\Delta}$) insertion, and order-based placement strategies (frontmost/middle position).
Each approach preserves different aspects of the original token relationships and context.
}
\label{fig:token arrangement}
\end{figure}

As illustrated in \cref{fig:token arrangement}, we comprehensively analyze various strategies for token arrangement after merging. 
In our example, using dst token 8 as an anchor, tokens 3, 5, and 7 are selected and 3, 5, and 7 with dst token 8 are merged into a single token 'a' through bipartite matching. 
We then explore different arrangement strategies for this merged token.

\noindent\textbf{(1) Isolated Placement.}
Isolated placement positions the merged tokens at the boundaries of the sequence:
\begin{itemize}
    \item Front placement: The merged token 'a' is positioned at the beginning of the sequence (position 0), followed by all unmerged tokens (1, 2, 4, 6, 9). 
    \item Last placement: The merged token 'a' is placed at the end of the sequence after all unmerged tokens. 

\end{itemize}

\noindent\textbf{(2) Internal Insertion.}
Internal insertion merging includes strategies for placing merged tokens at various positions within the sequence:

\noindent\textbf{(2.1) Destination Position.}
\begin{itemize}
    \item
    The merged token 'a' is positioned at the location of the destination token (position 8 in the example). 
    \end{itemize}
\textbf{(2.2) Importance-based Insertion.}
This strategy places merged tokens based on the informativeness ($\mathbf{\Delta}$) of the tokens being merged:
\begin{itemize}
\item
High $\mathbf{\Delta}$ placement: The merged token 'a' is positioned at the location of the most informative token among those being merged. In the example, token 5 has the highest informativeness value, so 'a' is placed at position 5.
\end{itemize}

\noindent\textbf{(2.3) Order-based Insertion.}
This approach uses the original sequential positions of the merged tokens to determine placement:
\begin{itemize}
\item Frontmost order position: The merged token 'a' is placed at the position of the frontmost token being merged (position 3 in the example). 
\item Mid-order position: The merged token 'a' is positioned at the middle token's location among those being merged (position 5 in the example). 
\end{itemize}
\label{sec:token arrangement}

\pagebreak
\clearpage
\subsection{Implementation Details.}
All experiments were conducted on four NVIDIA RTX 3090 GPUs, with video recognition tasks utilizing four NVIDIA A6000 GPUs. 
For computational efficiency measurements, we calculated FLOPs using the fvcore \footnote{https://github.com/facebookresearch/fvcore} library on a single RTX 3090.
For non-supported CUDA kernel operations (such as selective scan), we calculate it based on the previous advances\footnote{https://github.com/state-spaces/mamba/issues/110} considering its directionality.

\noindent\textbf{Model Configurations.} For ViM models, we applied token merging at layers 8, 14, and 20 with a reduction ratio $r=50$. 
MambaVision implementations merged tokens at stage 3's initial layer ($r=40$). we note that MambaVision requires additional sequence length restoration padding at the last stage due to architectural constraints.

\noindent\textbf{Training Details.} ViM models were trained with batch size 128, learning rate 7e-5 (cosine decay), weight decay 1e-6, and 5-epoch warmup following most settings in  \cite{vim_zhu}. 
MambaVision used learning rate 5e-4 and weight decay 0.05 following \cite{hatamizadeh2024mambavision}
For baseline comparison, we implemented a similarity-only variant equivalent to ToMe, ensuring identical experimental conditions for fair evaluation.

\section{Extended Results}
\label{Extended}

\subsection{Image Classification Results without Training}
\label{exp: without training}
\newcolumntype{Y}{>{\centering\arraybackslash}X}
\begin{table}[h]
\centering
\begin{tabularx}{0.47\textwidth}{c|YYYYY}
\toprule
$r$ & 1 & 5 & 10 & 30 & 50 \\
\midrule
\rowcolor{Gray}
& \multicolumn{5}{c}{Tiny} \\
ToMe  & 20.9 & 21.2 & 21.7 & 21.9 & 31.9 \\
ToMe (Ord.) & 76.4 & 76.2 & 76.0 & 73.8 & 66.8 \\
Ours      & 76.4 & 76.2 & 76.1& 74.6 &  67.2  \\
\midrule
FLOPs (G)    & 1.60 & 1.56 & 1.52 & 1.34 & 1.16 \\
\midrule
\rowcolor{Gray}
& \multicolumn{5}{c}{Small} \\
ToMe  & 73.7 & 73.9 & 73.8 & 72.7 & 73.8\\
ToMe (Ord.) & 80.1   & 80.0 & 80.0  & 79.6 & 74.1\\
Ours      & 80.1 & 80.1 & 80.1 & 79.8 & 74.6\\
\midrule
FLOPs (G)    & 5.28 & 5.16 & 5.01 & 4.41 & 3.81 \\

\bottomrule
\end{tabularx}
\caption{ImageNet-1K results without training. 
Tiny and Small denote ViM-T and ViM-S. $r$ denotes a number of reduced tokens, and ordered indicates token arrangement to original order.}
\label{tab:zero_shot_all}
\end{table}

The inference results without any additional training for both ViM-T and ViM-S models are presented in \cref{tab:zero_shot_all}.
The performance was evaluated by varying the number of reduced tokens from 1 to 50. \pagebreak
To comprehensively assess the contribution of each component, we designed a series of ablation experiments.

Our proposed method demonstrates superior performance compared to the ToMe without any additional training. 
For ViM-T with $r=50$, our method improves accuracy from 31.9\% to 67.2\%, while ViM-S shows an increase from 73.8\% to 74.6\%. 
In terms of computational efficiency, our method shows consistent reduction in computational costs as $r$ increases from 1 to 50, with FLOPs decreasing from 1.60G to 1.16G for ViM-T and from 5.28G to 3.81G for ViM-S.
While increasing the value of $r$ leads to slight performance trade-offs, our method still demonstrates substantial improvements over the ToMe approach, particularly for ViM-T.
\\

\subsection{Finetuning on High-resolution Images.}
\label{exp: higher-resolution}
\begin{table}[h]
\centering
\begin{tabular}{
>{\arraybackslash}p{0.2\columnwidth}|
>{\centering\arraybackslash}p{0.1\columnwidth}
>{\centering\arraybackslash}p{0.1\columnwidth}
>{\centering\arraybackslash}p{0.25\columnwidth}}

\toprule
Resolution & acc & gflops & ft time(4gpu)\\ 
\midrule
\textcolor{lightgray}{224$^2$} &\textcolor{lightgray}{76.1} &\textcolor{lightgray}{1.60}  
&\textcolor{lightgray}{-} \\
384$^2$ & 79.5 & 4.69 & 28 hrs\\
512$^2$ & 80.2 & 8.33 &  112 hrs$^\ddagger$ \\
\rowcolor{Gray}
224$^2_{r=50}$ & 76.2 & 1.14 & 8 hrs \\
\rowcolor{Gray}
384$^2_{r=150}$ & 78.9 & 3.33 &  20 hrs\\
\rowcolor{Gray}
512$^2_{r=250}$ & 79.8 & 6.10 & 61 hrs \\

\bottomrule
\end{tabular}
\caption{Finetuning on high-resolution images.
The light gray text indicates the performance of the base-model for comparison. 
$\ddagger$ indicates estimated results.}
\label{tab:high_res}
\end{table}

To evaluate our model on high-resolution images, as shown in \cref{tab:high_res}, we finetuned our model with varying image sizes: $224 \times 224$, $384 \times 384$, and $512 \times 512$. 
Since high resolutions generate more tokens (e.g., 1025 tokens for $512^2$ with patch size 16), we adjusted the number of reduced tokens $r$ to 50, 150, and 250 tokens, respectively.
Our method maintains competitive accuracy while reducing computational cost by around 30\% and training time by up to 46\%.

\clearpage

\section{Additional Visualizations} \label{exp: vis}

\begin{minipage}[t]{\textwidth}
    \includegraphics[width=\linewidth]{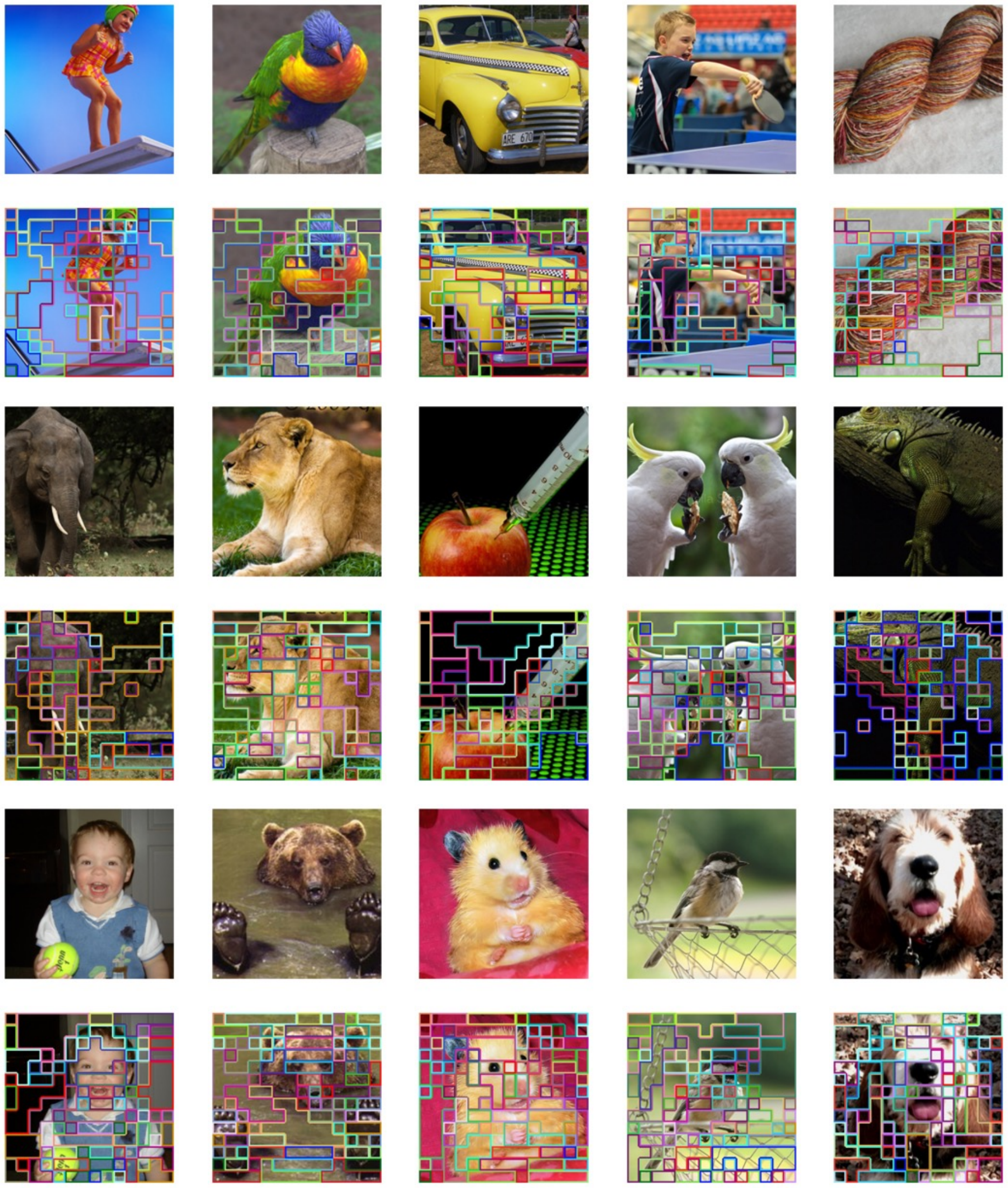}
    \captionof{figure}{Example pairs of input images (top) and MaMe's token-merging results (bottom). Patches with the same border color represent tokens that have been merged together.}
    \label{fig:additional_visualizations}
\end{minipage}

\end{document}